\pdfoutput=1

\documentclass[11pt]{article}

\usepackage[]{acl}
\usepackage{mainlp} 

\usepackage{wrapfig}
\definecolor{light-gray}{gray}{0.95}
\usepackage{tcolorbox}

\usepackage{microtype}

\usepackage{caption}
\usepackage{subcaption}
\usepackage{listings}
\usepackage{color}
\usepackage{soul}
\definecolor{lightgrey}{rgb}{0.925, 0.925, 0.925}
\sethlcolor{lightgrey}

\usepackage{todonotes}
\usepackage{booktabs}

\usepackage{hyperref}
\newcommand{\corpus}[0]{\textsc{ClimatELi}}

\newcommand{\entitylink}[2]{\href{#2}{\textcolor{black}{\texttt{#1}}}}

\newcommand{\yes}{\ding{51}}

\title{\corpus{}: Evaluating Entity Linking on Climate Change Data} 

\author{Shijia Zhou\thanks{\;\; Equal contribution.}
\quad \quad
Siyao Peng\footnotemark[1]
\quad \quad
Barbara Plank\\
MaiNLP, Center for Information and Language Processing, LMU Munich, Germany \\
Munich Center for Machine Learning (MCML), Munich, Germany \\
{\tt zhou.shijia@campus.lmu.de \{siyao.peng,b.plank\}@lmu.de}}

\begin{document}
\maketitle

\begin{abstract}
Climate Change (CC) is a pressing topic of global importance, attracting increasing attention across research fields, from social sciences to Natural Language Processing (NLP).
CC is also discussed in various settings and communication platforms, from academic publications to social media forums. 
Understanding who and what is mentioned in such data is a first critical step to gaining new insights into CC.
We present \corpus{} (\textsc{Climat}e Entity \textsc{Li}nking), the first manually annotated CC dataset that links 3,087 entity spans to Wikipedia. 
Using \corpus{}, we evaluate existing entity linking (EL) systems on the CC topic across various genres and propose automated filtering methods for CC entities.
We find that the performance of EL models notably lags behind humans at both token and entity levels.
Testing within the scope of retaining or excluding non-nominal and/or non-CC entities particularly impacts the models' performances.
 \end{abstract}

\section{Introduction}\label{sec:introduction}

Climate change (CC) is a well-established and omnipresent concept influencing daily lives.
Natural Language Processing (NLP) tasks such as Entity linking (EL) facilitate knowledge base population and empower individuals to enhance their CC understanding. 
For example, after reading the news about ``young indigenous women from Mexico and Morocco unite for COP27,''
one could click the following Wikipedia links and explore further about \entitylink{COP27}{https://en.wikipedia.org/wiki/2022_United_Nations_Climate_Change_Conference}
or \entitylink{Climate\_of\_Mexico}{https://en.wikipedia.org/wiki/Climate_of_Mexico}.
EL provides easy access to CC-related knowledge and endows downstream applications like extracting information regarding stakeholders: policymakers, scientists, administrators, and etc. \citep{conde2005engaging}. 

\begin{figure}
\centering
\begin{tcolorbox}[colback=white, colframe=black, fonttitle=\bfseries, coltitle=black, colbacktitle=white, boxrule=0.5mm, left=2mm, right=2mm, top=2mm, bottom=2mm]
\normalsize
According to the [World Meteorological Orga-\\nization](\href{http://en.wikipedia.org/wiki/World_Meteorological_Organization}{http://en.wikipedia.org/wiki/World\_\\Meteorological\_Organization}), [La Niña](\href{https://en.wikipedia.org/wiki/El_Ni%C3%B1o%E2%80%93Southern_Oscillation}{El\_\\Niño-Southern\_Oscillation}) ``refers to the large-scale [cooling](\href{http://en.wikipedia.org/wiki/Heat_transfer}{Heat\_transfer}) of [the ocean surface temperatures](\href{http://en.wikipedia.org/wiki/Sea_surface_temperature}{Sea\_surface\_temperature}) in the central 
and eastern [equatorial](\href{http://en.wikipedia.org/wiki/Equator}{Equator}) 
[Pacific Ocean](\href{http://en.wikipedia.org/wiki/Equator}{Pacific\_Ocean})''.

\end{tcolorbox}
\caption{Sample \corpus{} annotation.}
\label{fig:enter-label}
\end{figure}

To gain valuable and broad access to relevant CC information, a major challenge is that EL needs to robustly process different texts, both in terms of the specific CC topic and across a wide range of genres. However, there is a lack of human-labeled evaluation data for EL on CC, so little is known about how well existing EL perform. 

We present \corpus{} \footnote{\url{https://github.com/mainlp/ClimatELi}}, the first manually annotated CC EL corpus covering five English genres in \S\ref{sec:corpus}. 
\S\ref{sec:entity-linking-evaluation} introduces three EL systems for evaluation and applies an automatic filtering mechanism for CC-related and nominal entities. 
\S\ref{sub:results}-\S\ref{sec:qualitative-analysis} analyze models' overall performance and on specific genres and threshold conditions.
\S\ref{sec:conclusion} concludes the paper.

\section{Related Work}\label{sec:related-work}

Entity-level information is gaining increasing attention in CC-related research.
Different from named entity recognition (NER) applications on CC texts \citep{maynard_tweet_2015, mishra2021neuralnere, piskorski_exploring_2022, vaid_towards_2022, spezzatti-sdg-2022}, entity linkers (EL) further disambiguate textual mentions and associate them to knowledge bases (KBs), 
e.g., Wikipedia \citep{cucerzan-2007-large},
DBPedia \citep{mendes-2011-dbpedia, mendes-etal-2012-dbpedia},  Wikidata \citep{vrandevcic2014wikidata}, via candidate selection and ranking \citep{rao_linking_2013, hachey-evaluating-2013, moro-etal-2014-entity, shen-linking-2015}.
EL evaluations also spread from news, e.g., TAC-KBP \cite{mcnamee2009overview, ji2010overview}, to more genres \citep{derczynski-linking-2015, yang-chang-2015-mart, lin-zeldes-2021-wikigum}, and topic domains \citep{klie-etal-2020-zero, liu-etal-2021-learning-domain}.

Despite a lack of human evaluation benchmarks for EL on CC data, there is work on integrating EL as an automatic pipeline for information extraction. 
\citet{weichselbraun-consolidating-2015} present Recognyze, an EL platform confined to named entities adopted by two CC web applications on Swiss business news, i.e., the Media Watch on Climate Change and the Climate Resilience Toolkit. 
\citet{ruiz-etal-2016-word} propose an NLP pipeline including EL for identifying supporting and opposing propositions in CC data.
\citet{diggelmann_climate-fever_2021} employ EL to extract relevant documents for evidence candidates in CC claim verification.
\citet{perezortiz-watch-2022} conduct EL on video transcriptions surrounding two topics---``machine learning'' and ``climate change''---to speed up video search. 
\citet{toulet-issa-2022} provide an EL-integrated pipeline 
and a visualization tool 
for analyzing scientific articles, including a case study on co-occurring ``climate change'' and ``health'' topics.
\citet{pitacosta-towards-2024} rely on EL as a semantic annotation to analyze the impact of water-related climatic disasters.
However, no evaluation of EL performance on English CC data across genres exists.

\section{\corpus{}: \textsc{Climat}e Entity 
\textsc{Li}nking}\label{sec:corpus}

We present \corpus{}, the first EL evaluation corpus on CC data. 
We include ten English documents across five genres---Wikipedia pages, academic articles \footnote{\url{https://www.mdpi.com/}}, web news \footnote{\url{https://globalvoices.org/}}, United Nations' Intergovernmental Panel on Climate Change reports\footnote{\url{https://www.ipcc.ch/report/ar6/wg3/downloads/report/IPCC_AR6_WGIII_FullReport.pdf}}, and YouTube transcriptions---consisting of 3,087 (1,161 unique) entity links across 12,802 tokens. 

\paragraph{Entity Linking}
Manual EL annotation from scratch is time-consuming and might result in low recall.
This is mostly due to annotators' inability to encompass all contents from ever-expanding knowledge bases (KBs). 
We pick Wikipedia as our target knowledge base due to its enormous size\footnote{English Wikipedia contains 6.8M articles as of June 20, 2024, see \entitylink{Wikipedia:Size\_of\_Wikipedia}{https://en.wikipedia.org/wiki/Wikipedia:Size_of_Wikipedia}.} and its integration into our daily life. 

\begin{table}[t]
\centering
\resizebox{\columnwidth}{!}{
\begin{tabular}{c|cc|ccc}
\toprule
& \multicolumn{2}{c|}{token-level}          & \multicolumn{3}{c}{entity-level}   \\ 
& accuracy & cohen's $\kappa$ & precision & recall & F1 \\
\midrule
untyped & 94.31 & 89.93 & 89.90 & 82.63 & 86.12  \\
typed & 92.85 & 88.94 & 87.30 & 80.24 & 83.62  \\
\bottomrule
\end{tabular}
} 
\caption{\corpus{} inter-annotator agreement. }
\label{tab:IAA}
\end{table}

Firstly, we manually correct tokenization and POS-tagging from stanza \citep{qi-etal-2020-stanza} predictions. 
We then create \corpus{} by opting to manually correct pre-tagged linked entities from Wikifier's \citep{brank-wikifier-2017} threshold 1.0 (highest recall) predictions, which provide an extensive baseline for gold annotations while also adding missing entities.
We use the markdown style \texttt{[Document Tokens](\entitylink{Link\_URL}{http://en.wikipedia.org/wiki/})} to ease annotation as shown in Figure \ref{fig:enter-label}.
Annotations include removing or correcting wrongly detected entity spans and links predicted by Wikifier and adding missing entities not annotated by Wikifier. 
We also verify whether individual links exist on Wikipedia and resolve various ambiguities.

We next also discuss several \corpus{} guidelines. Firstly, we follow the flat schema of Wikipedia webpages to annotate only the longest entity without the shorter nested ones. 
For example, if ``climate change mitigation'' appears, we only link the entirety to \entitylink{Climate\_change\_mitigation}{https://en.wikipedia.org/wiki/Climate_change_mitigation}, without annotating the shorter \entitylink{Climate\_change}{https://en.wikipedia.org/wiki/Climate_change}  or \entitylink{Climate}{https://en.wikipedia.org/wiki/Climate}.
Secondly, we determine the associated Wikipedia link based on an entity's contextual reading.
For example, when ``it'' refers to ``climate change'', we annotate ``[it's effect](\entitylink{Effects\_of\_Climate\_Change}{https://en.wikipedia.org/wiki/Effects_of_climate_change})''.
Similarly, if the Kyoto Protocol is in context, we annotate the subsequent ``[the protocol](\entitylink{Kyoto\_Protocol}{http://en.wikipedia.org/wiki/Kyoto_Protocol}).''
Lastly, since EL KBs are not tailored for CC, we follow Wikipedia to annotate ELs on generic verbs, adjectives, adverbs, conjunctions, and etc., if such entries exist. 
These include 
``refer'' (a verbal form of \entitylink{Reference}{https://en.wikipedia.org/wiki/Reference}), 
``possible'' (an adjectival form of \entitylink{Logical\_possibility}{http://en.wikipedia.org/wiki/Logical_possibility}), 
``successfully'' (an adverbial form of \entitylink{Success}{https://en.wikipedia.org/wiki/Success}), 
``while'' (\entitylink{While}{https://en.wikipedia.org/wiki/While}), and etc.

\begin{table*}[t]
\centering
\resizebox{0.85\textwidth}{!}{
\begin{tabular}{c|cccc|rr|rr|rr|rr}
\toprule
\multirow{2}{*}{\begin{tabular}[c]{@{}l@{}}Combos \end{tabular}}
& \multicolumn{4}{c|}{Filters} 
 & \multicolumn{2}{c|}{Gold}
& \multicolumn{2}{c|}{Wikifier}
& \multicolumn{2}{c|}{TagMe}
& \multicolumn{2}{c}{\citeauthor{cao2021autoregressive}}
\\
 & \textit{Valid} & \textit{Nom} & \textit{CText} & \textit{CLink} 
& Total & Unique
& Total & Unique
& Total & Unique
& Total & Unique
\\
\midrule
\textit{Orig} & &  &  &  &   3,087 & 1,161  &  4,823 & 1,730  &  4,165 & 2,129  &  1,399 & 576   \\
/ & \yes & &  &  &   3,061 & 1,141  &  4,783 & 1,714  &  4,072 & 2,082  &  1,304 & 511   \\
\textit{N-only} & \yes &\yes & &  &   2,346 & 881  &  2,587 & 1,008  &  2,764 & 1,333  &  1,106 & 437   \\
\textit{C-only} & \yes & & \yes & &   1,831 & 558  &  2,030 & 605  &  1,875 & 683  &  977 & 338   \\
/ & \yes & &  & \yes &  958 & 230  &  1,009 & 211  &  855 & 220  &  479 & 139   \\
\textit{NC-only} & \yes &\yes &\yes & &   1,586 & 491  &  1,554 & 481  &  1,557 & 565  &  896 & 316   \\
/ & \yes &\yes & & \yes &  872 & 209  &  830 & 185  &  772 & 197  &  457 & 129   \\
\bottomrule
\end{tabular}
}
\caption{Frequencies of all versus unique entity links in human and model annotations under different filters.}
\label{tab:entity_link_stats_filtering}
\end{table*}

\paragraph{Inter-Annotator Agreement}
Two authors of this paper, fluent English speakers, split the ten documents, each taking $\sim$3 hours to annotate. 
We evaluate inter-annotator agreement (IAA) on one Wikipedia article, \entitylink{Paris\_Agreement}{https://en.wikipedia.org/wiki/Paris_Agreement}, which includes 1,371 tokens, and 334 or 307 entity link annotations by two annotators, respectively. 
Table \ref{tab:IAA} presents our IAA using accuracy, Cohen's kappa at the token level, precision, recall, and F1 at the entity level.
We also include both untyped and typed results, where the former only matches the entity span, and the latter requires annotating the same Wikipedia link. 
We achieve high  (80\%+) token and entity-level IAAs with precision scoring higher than recall on both untyped and typed entities. 

\begin{table*}[t]
\centering
\resizebox{0.70\textwidth}{!}{
\begin{tabular}{ll|c|ccc|c|ccc}
\toprule
\multirow{2}{*}{Combos} &  \multirow{2}{*}{Models} & \multicolumn{4}{c|}{untyped}     & \multicolumn{4}{c}{typed}   \\ 
&  & accuracy & precision & recall& F1 & accuracy & precision & recall & F1 \\ \midrule
\multirow{3}{*}{\textit{Orig}} & Wikifier & \textbf{76.02} & 49.55 & \textbf{77.42} & \textbf{60.43} & 68.64 & 38.36 & \textbf{59.93} & \textbf{46.78}  \\
& Tagme &	75.39&	47.71&	64.37&	54.80 & 57.03& 16.47&	22.22&	18.92 \\
& \citeauthor{cao2021autoregressive} & 72.98 & \textbf{55.90}  & 25.33 & 34.86  & \textbf{70.15} & \textbf{45.46} & 20.60  & 28.35 \\
\midrule
\multirow{3}{*}{\textit{N-only}} & Wikifier  & \textbf{87.95} & \textbf{68.07} & \textbf{75.06} & \textbf{71.40} & \textbf{82.67} & 52.34 & \textbf{57.72} & \textbf{54.90}\\
& Tagme & 84.82 & 57.20 & 67.39 & 61.88  & 70.30 & 20.12 & 23.70 & 21.76 \\
& \citeauthor{cao2021autoregressive} & 79.94 & 63.56 & 29.97 & 40.73 & 77.67 & \textbf{53.35} & 25.15 & 34.18  \\
\midrule
\multirow{3}{*}{\textit{C-only}}
& Wikifier & \textbf{87.92} & \textbf{60.20} & \textbf{66.74} & \textbf{63.30} & \textbf{84.96} & \textbf{51.63} & \textbf{57.24} & \textbf{54.29} \\
& Tagme & 86.73 & 54.35 & 55.65 & 54.99 & 76.61 & 22.13 & 22.67 & 22.40\\
& \citeauthor{cao2021autoregressive} & 84.12 & 58.34 & 31.13 & 40.60 & 82.14 & 50.56 & 26.98 & 35.19 \\
\midrule
\multirow{3}{*}{\textit{NC-only}}
& Wikifier & \textbf{90.27} & \textbf{68.02} & \textbf{66.65} & \textbf{67.32}  & \textbf{87.62} & \textbf{57.79} & \textbf{56.62} & \textbf{57.20} \\ 
& Tagme & 88.88 & 58.83 & 57.76 & 58.29 & 79.43 & 23.31 & 22.89 & 23.10 \\
& \citeauthor{cao2021autoregressive} & 85.92 & 60.71 & 34.30 & 43.84 & 84.09 & 52.90 & 29.89 & 38.20 \\
\bottomrule
\end{tabular}
}
\caption{Typed and untyped token-level accuracy and entity-level  precision, recall, and F1
scores.
}
\label{tab:without_threshold}
\end{table*}

\section{Evaluation Setup}\label{sec:entity-linking-evaluation}

This section presents the experiment setups for evaluating entity-linking models on \corpus{} data.

\paragraph{Entity Linkers}
We employ three Wikipedia linking models frequently used in NLP and social science to generate EL predictions. 
Wikifier \citep{brank-wikifier-2017} and TagMe \citep{ferragina-tagme-2010} are easy to use and allow users to configure the confidence thresholds on predicted entity links. 
Additionally, we include a BART-based \citep{lewis-etal-2020-bart} generative sequence-to-sequence EL model, \citet{cao2021autoregressive}.\footnote{We use \citeauthor{cao2021autoregressive} rather than GENRE 
(Generative ENtity REtrieval) to refer to the third EL model to avoid confusion between the model and \corpus{}'s text genres.}
Since entity spans differ vastly across ELs, we conduct a post-processing step to normalize predicted entities: removing leading determiners from nouns and dropping nested entities following Wikipedia's style, which displays no nested or overlapping ELs.

\paragraph{Entity Filters 
}\label{subsec:entity-filter}
We design filters to focus our EL evaluation on valid, nominal, and climate entities. 

\begin{itemize}[leftmargin=*] 
\setlength\itemsep{0.5em}

\item \textit{Valid links (Valid)}: 
we discard disambiguation and content-less pages, e.g., \entitylink{Reduction}{https://en.wikipedia.org/wiki/Reduction} and \entitylink{Climate\_overshoot}{https://en.wikipedia.org/wiki/Climate_overshoot}, 
and invalid URLs;

\item \textit{Nominal (Nom)}:
Since ELs such as Wikifier provide links to verbs, e.g., \textit{`he [thinks](\entitylink{Thought}{https://en.wikipedia.org/wiki/Thought}) ...'}, we 
remove non-nominal entities whose entirety are verbs, adjectives, adverbs, etc.;

\item  \textit{Climate Text or Link}: 
we only retain Wikipedia links that either include the word ``climate'' (\textit{CText}) or the link \entitylink{Climate\_change}{https://en.wikipedia.org/wiki/Climate_change} (\textit{CLink}).
\end{itemize}

Table \ref{tab:entity_link_stats_filtering} presents the frequency of linked entities and unique ones from gold human annotations and model predictions in different filtering scenarios. 
We observe that \textit{Nom} reduces valid gold entities from 3,061 to 2,346, \textit{CText} to 1,831, and \textit{CLink} to 958.
The number of unique gold entities is halved after \textit{CText}-filtering and quartered by \textit{CLink}, and drops more dramatically on model predictions. 
Since \textit{CLink} is more restrictive and has a lower recall than \textit{CText}, we use \textit{CText} as the climate filter in evaluations. 
Moreover, Wikifier and TagMe generate more predictions than humans, whereas \citeauthor{cao2021autoregressive} is more conservative. 
In addition to the annotations on 12K tokens, we release a list of 1,251 CC-related Wikipedia links for future research.

\section{Results}\label{sub:results}

This section evaluates four filtering scenarios, comparing \textit{Valid Nom} and \textit{Valid CText} versions (\textit{N-only} and \textit{C-only})  and their intersections (\textit{NC-only)} to unfiltered (\textit{Orig}) entities. 
For Wikifier and TagMe, we use the default confidence threshold with the highest recall. 
Table \ref{tab:without_threshold} presents the overall token- and entity-level performances on untyped and typed entities under four filtering scenarios.

\paragraph{Token-level}
As the filtering conditions become more stringent, the accuracy increases for all three models.
Namely, \textit{NC-only} achieves the highest accuracy, followed by \textit{C-only} and \textit{N-only}, and \textit{Orig} scores the lowest.
TagMe exhibits the largest disparity between typed and untyped token accuracy among the three models, but this difference decreases after adding filters.
Nevertheless, comparing token-level accuracy between unfiltered and filtered versions is unfair since the latter has more non-entity tokens and raises chance agreement. 

\begin{figure*}[t!bh]
\centering
\includegraphics[width=0.98\textwidth]{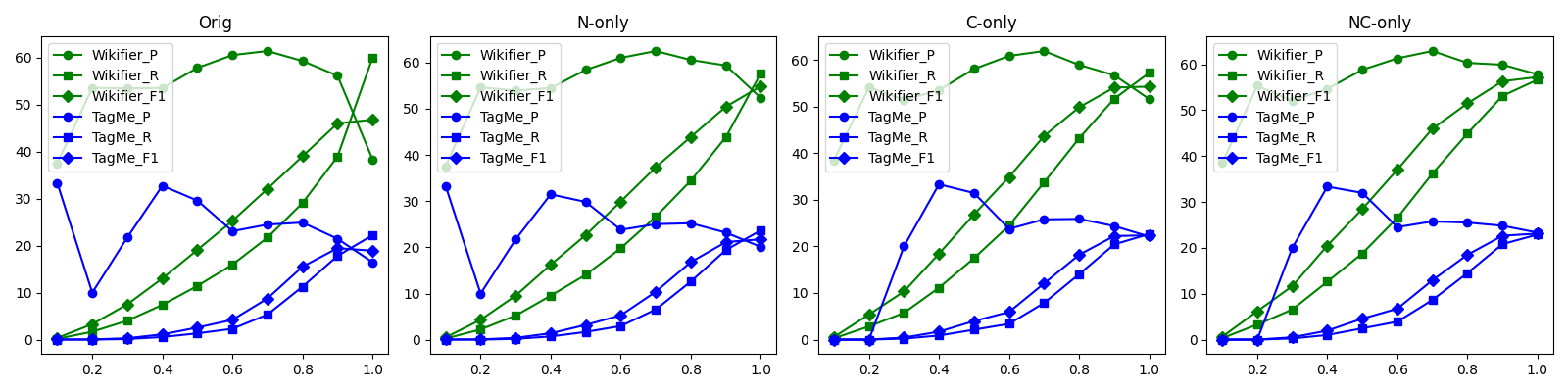}
\caption{Typed entity precision, recall, and F1 of Wikifier and TagMe on thresholds 0.1 to 1.0 (min to max recall).}
\label{fig:lineplot_threshold}
\end{figure*}

\paragraph{Entity-level} 
Entity-level results reveal difficulties in EL, with all typed F1s below 60\%.
Wikifier remains the winner for both untyped and typed F1s.
TagMe achieves satisfying performance on retrieving entity spans but deteriorates largely on typed scores.
\citeauthor{cao2021autoregressive} perform the worst on untyped scores, but its degradation to typed is relatively small. 
Besides, due to fewer predicted entities (cf. Table \ref{tab:entity_link_stats_filtering}), \citeauthor{cao2021autoregressive} demonstrates higher precision than recall, whereas TagMe and Wikifier prefer recall. 
However, on \textit{NC-only} entities, Wikifier and TagMe's privilege in recall diminishes, and all three models exhibit higher precision than recall.

\section{Quantitative Analysis}\label{sec:qualitative-analysis}
We examine two impacting factors on EL model scores: confidence thresholds and text genres.

\paragraph{Wikifier and TagMe Thresholds}\label{subsec:wikifier-threshold}
Figure \ref{fig:lineplot_threshold} visualizes how different Wikifier and TagMe confidence thresholds affect model performances.
Unsurprisingly, Wikifier beats TagMe on all thresholds. On the more filtered data, the recall of both models increases with the threshold. 
In contrast, precision fluctuates: Wikifier dips around 0.2 and peaks around 0.7; TagMe reaches its best precision around 0.4 and decreases thereafter.
Moreover, it is intriguing that the precision and recall of both models converge at threshold 1.0  under the \textit{NC-only} filter, which means the classifiers are balanced in terms of sensitivity and specificity.

\paragraph{Genres}\label{subsec:genre}
\corpus{} contains documents from 5 genres with different linguistic styles and discourse structures. 
Given our interest in CC, Table \ref{tab:5_genres_nc} presents genre performances on \textit{NC-only} data. 
All three models achieve the highest performance in the vlog genre.
Though TagMe and \citeauthor{cao2021autoregressive} perform inferior to Wikifier, their advantages are remarkable in vlogs compared to other genres.

\begin{figure}[t]
\centering
\includegraphics[width = 0.49\textwidth]{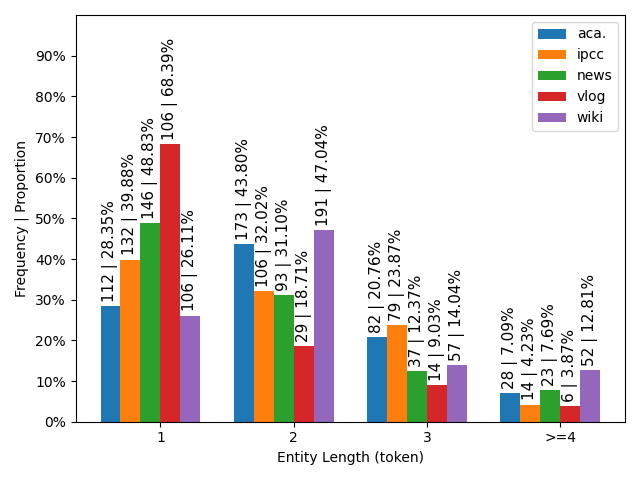}
\caption{\textit{NC-only} entity length distributions per genre.}
\label{fig:NC_length_distribution}
\end{figure}

Figure \ref{fig:NC_length_distribution} visualizes the distribution of entity lengths across various genres and illustrates that most (68.39\%) entity spans in vlogs include only one token. This reduces the likelihood of entity nesting and decreases the difficulty of determining the positions of entity spans.
Meanwhile, academic, ipcc, and wiki have considerably longer entities, possibly explaining TagMe and \citeauthor{cao2021autoregressive}' sub-optimal performances in these three genres.

\section{Error Types}\label{sec:error-types}

This section further exemplifies common error types of the EL models' predictions.

\paragraph{Misinterpreting polysemous words as false nominals}
Polysemous tokens occurring in non-nominal positions were sometimes wrongly linked to their nominal interpretations.
For example, the adjective ``current'' (i.e., belonging to the present time)  receives a wrong link to \entitylink{Ocean\_current}{http://en.wikipedia.org/wiki/Ocean_current};
the coordinating conjunction ``both'' gets wrongly linked to a Trap song \entitylink{Both\_(song)}{http://en.wikipedia.org/wiki/Both_(song)};
the 
modal verb ``will'' gets falsely associated with
\entitylink{Free\_will}{http://en.wikipedia.org/wiki/Free_will} (i.e., capacity to make decisions independently).

\paragraph{Overly specifying generic nouns to particular readings}
Entity linkers, particularly Wikifier with threshold 1.0, are inclined to magnify the specificity of nominal terms and impose or enforce an association to some Wikipedia page. 
The ``meeting'' of governmental parties is distorted to 
\entitylink{Confluence}{http://en.wikipedia.org/wiki/Confluence} (i.e., the joining of two watercourses into one);
``organizations'' in general gets over-specified to \entitylink{Non-governmental\_organization}{http://en.wikipedia.org/wiki/Non-governmental_organization}.
Particularly when the more general or common interpretation of a noun is absent in Wikipedia,
models tend to impose an association with an overly specified EL. 
For example, a ``step'' as part of a plan is misrepresented as a 
\entitylink{Step\_dance}{http://en.wikipedia.org/wiki/Step_dance}, 
a stair step
(\entitylink{Stairs}{http://en.wikipedia.org/wiki/Stairs}),
or the mathematical 
\entitylink{Step\_function}{https://en.wikipedia.org/wiki/Step_function}.
Similarly, most ``growth''s are forcibly linked to
\entitylink{Population\_growth}{https://en.wikipedia.org/wiki/Population_growth}
or 
\entitylink{Economic\_growth}{https://en.wikipedia.org/wiki/Economic_growth}
even when referring to other objects' increase in size.

\begin{table}[t]
\centering
\resizebox{0.35\textwidth}{!}{
\begin{tabular}{ll|ccc}
\toprule
Models & Genre & precision & recall & F1  \\ \midrule
\multirow{5}{*}{Wikifier} & aca.             & 58.35 & 57.47 & 57.91 \\
& ipcc                 & 55.93 & 55.59 & 55.76 \\
& news                 & 57.77 & 57.19 & 57.48 \\
& vlog                 & \textbf{64.85} & \textbf{69.03} & \textbf{66.88} \\
& wiki                 & 55.73 & 51.48 & 53.52 \\ \midrule
\multirow{5}{*}{TagMe}    & aca.             & 19.95 & 20.51 & 20.22 \\
& ipcc                 & 24.84 & 23.56 & 24.19 \\
& news                 & 26.51 & 26.42 & 26.47 \\
& vlog                 & \textbf{43.71} & \textbf{42.58} & \textbf{43.14} \\
& wiki                 & 15.21 & 14.53 & 14.86 \\ \midrule
\multirow{5}{*}{\citeauthor{cao2021autoregressive}}    & aca.             & 47.26 & 24.05 & 31.88 \\
& ipcc                 & 51.20 & 19.34 & 28.07 \\
& news                 & 61.02 & 36.12 & 45.38 \\
& vlog                 & \textbf{66.40} & \textbf{53.55} & \textbf{59.29} \\
& wiki                 & 46.27 & 30.54 & 36.80 \\ 
\bottomrule
\end{tabular} 
}
\caption{Typed entity scores on
5 \textit{NC-only} genres.}
\label{tab:5_genres_nc}
\end{table}

\paragraph{Unable to capture contextual readings}
In some cases, models succeed in annotating ``[its effects]\entitylink{(Effects\_of\_climate\_change)}{http://en.wikipedia.org/wiki/Effects_of_climate_change}''
when the pronoun ``it'' refers to 
\entitylink{Climate\_change}{https://en.wikipedia.org/wiki/Climate_change} or 
``Both [the EU](\entitylink{European\_Union}{http://en.wikipedia.org/wiki/European_Union})
and [its member states](\entitylink{Member\_state\_of\_the\_European\_Union}{http://en.wikipedia.org/wiki/Member_state_of_the_European_Union})'' when ``it'' refers \entitylink{European\_Union}{http://en.wikipedia.org/wiki/European_Union}.
However, the coreference of ``the''-headed definite common nouns is difficult. 
For example, when ``the Accord'' refers to \entitylink{Copenhagen\_Accord}{https://en.wikipedia.org/wiki/Copenhagen_Accord},
it gets falsely linked to \entitylink{Prices\_and\_Incomes\_Accord}{http://en.wikipedia.org/wiki/Prices_and_Incomes_Accord} in Australia.
Similarly, a coreferring ``the agreement'' to \entitylink{Paris\_Agreement}{https://en.wikipedia.org/wiki/Paris_Agreement} is always mis-interpreted as 
\entitylink{Joint\_Comprehensive\_Plan\_of\_Action}{http://en.wikipedia.org/wiki/Joint_Comprehensive_Plan_of_Action}, i.e., the Iran Nuclear Deal.  
The more tricky situation is when some ELs occur nearby but are irrelevant to the current entity. 
Still, EL models are falsely influenced by these contexts. 
In ``the 195 UNFCCC participating member states and the European Union'', TagMe annotates ``member states'' wrongly as \entitylink{Member\_state\_of\_the\_European\_Union}{http://en.wikipedia.org/wiki/Member\_state\_of\_the\_European\_Union}, and \\ Wikifier annotates them as \entitylink{European\_Union}{http://en.wikipedia.org/wiki/European\_Union}, while the gold label should be \entitylink{List\_of\_parties\_to\_\\the\_United\_Nations\_Framework\_Convention\_\\on\_Climate\_Change}{http://en.wikipedia.org/wiki/List\_of\_parties\_to\_the\_United\_Nations\_Framework\_Convention\_on\_Climate\_Change}. 

\paragraph{Ambiguity caused by the flat schema} 
Annotating the longest entity and ignoring the nested ones is common in named entity linking and recognition \citep{mcnamee2009overview, ji2010overview, tjong-kim-sang-de-meulder-2003-introduction}.
However, this results in human label variation in annotating overlapping ELs. 
For example, ``adapt to climate change effects'' can be annotated as ``[adapt to climate change effects](\entitylink{Climate\_change\_adaptation}{http://en.wikipedia.org/wiki/Climate\_change\_adaptation})'', but can also be annotated as ``[adapt to climate change](\entitylink{Climate\_change\_adaptation}{http://en.wikipedia.org/wiki/Climate\_change\_adaptation}) [effects](\entitylink{Effects\_of\_climate\_change}{http://en.wikipedia.org/wiki/Effects\_of\_climate\_change})''. 
Both interpretations are equally reasonable and converge with the annotation guidelines.

\section{Conclusion}\label{sec:conclusion}

This paper presents \corpus{}, an entity linking corpus for English climate change data on five genres. 
We analyze existing EL systems and propose filters to focus the evaluation on nominal climate entities. 
We show that ELs struggle to detect long entity spans and link them to corresponding Wikipedia pages.
Future work includes nested EL annotations since CC-related terminologies exhibit nesting and overlap, expanding annotated texts and Wikipedia links to more languages, and training a CC-adapted EL model for downstream NLP tasks.

\section*{Limitation}
There are a few limitations in our work that we plan to improve in future research. 
Firstly, we only annotated flat entity linking without nesting; thus, embedded ELs and co-occurrences between nested entities are not fully captured. 
Secondly, our annotated texts and Wikipedia links are limited to English, and we only evaluate the performances of English EL models.
Future expansion of multilingual texts and Wikipedia entries would benefit cross-lingual and cross-national comparison studies.
Thirdly, although EL models benefit from being domain-generic, researchers are interested in evaluating them on specific domains and could follow different logical approaches.
This paper employs simple rule-based filtering on manual annotations and model predictions to assess EL on CC-related data. 
However, with more CC-specific EL data available, we could finetune EL models on domain-specific data for direct evaluation.

\section*{Acknowledgements}
This work belongs to the KLIMA-MEMES project funded by the Bavarian Research Institute for Digital Transformation (bidt), an institute of the Bavarian Academy of Sciences and Humanities. The authors are responsible for the content of this publication.

\bibliography{main}

\end{document}